\documentclass[letterpaper,10pt,conference]{ieeeconf}
\usepackage[T1]{fontenc}
\usepackage[utf8]{inputenc}
\usepackage{cite}
\usepackage{amsmath,amssymb,bm}
\usepackage{booktabs,array,graphicx}
\usepackage{url}
\usepackage{todonotes}
\usepackage{float}
\usepackage{comment}
\usepackage[hidelinks]{hyperref}
\usepackage[normalem]{ulem}
\IEEEoverridecommandlockouts
\graphicspath{{Pictures/}}

\newif\ifanonymous
\anonymousfalse
\newcommand{\R}{\mathbb{R}}
\newcommand{\T}{\mathsf{T}}
\newcommand{\tr}{\operatorname{tr}}

\newcommand{\diag}{\operatorname{diag}}
\newcommand{\argmin}{\operatorname*{arg\,min}}
\newcommand{\norm}[1]{\left\lVert#1\right\rVert}
\newcommand{\Ap}{A_{p}}
\newcommand{\Au}{A_{u}}
\newcommand{\ez}{\bm{e}_{z}}
\hypersetup{pdftitle={Information-based Design Framework for Wireless NMR  Magnetic Tracking System in Medical Robots},pdfauthor={},pdfsubject={Covariance-based design of ultra-high-field magnetic tracking}}

\title{Information-based Unified Design Framework for \\ NMR Magnetic Tracking System}
\ifanonymous
  \author{\mbox{}\vspace{1em}}
\else
  \hypersetup{pdfauthor={Se\c{c}kin Eren Yetim and Mehmet Efe Tiryaki}}
  \author{Se\c{c}kin~Eren~Yetim$^{1}$ and Mehmet~Efe~Tiryaki$^{1,2}$%
  \thanks{$^{1}$The authors are with METU-ROMER, Center for Robotics and AI,
  Middle East Technical University, Ankara, T{\"u}rkiye.}%
  \thanks{$^{2}$Mehmet Efe Tiryaki is with the Department of Mechanical
  Engineering, Middle East Technical University, Ankara, T{\"u}rkiye.
  Corresponding author: etiryaki@metu.edu.tr.}}
\fi

\begin{document}
\bstctlcite{IEEEexample:BSTcontrol}
\maketitle
\thispagestyle{empty}
\pagestyle{empty}

\begin{abstract}
Nuclear magnetic resonance (NMR) magnetic tracking offers a promising alternative to magnetic resonance imaging (MRI)-based localization for medical robotic guidance. However, scalar field measurements and field-aligned marker moments constrain position and orientation tracking, requiring careful configuration design. We present a unified information-based framework that uses local covariance analysis to evaluate receiver geometry, marker separation, and external source configuration in on/off-board NMR tracking. Numerical evaluation shows that planar-array precision depends jointly on receiver count and geometry, with triangular lattice arrays achieving the lowest best-case position uncertainty among the evaluated layouts for 9--20 receivers. Two spatially separated, field-aligned markers enable device-axis tracking, with predicted angular uncertainty below $2^\circ$ at the evaluated workspace center for suitable marker separations. We further establish point-dipole source--receiver duality, unifying the analysis of on/off-board NMR tracking. These findings provide model-based guidance for tracking configuration design.
\end{abstract}
\vspace{-0pt}
\vspace{-12pt}

\section{Introduction}

NMR sensing has been demonstrated for patient head-motion tracking and wireless localization of medical tools in MRI scanners \cite{haeberlin2015motion,aranovitch2018prospective,tiryaki2025nmr}. This approach could provide an alternative to MRI-based localization for minimally invasive robotic interventions, including steerable guidewires and robotic catheters \cite{tiryaki2023magnetic,dong2022shape,murasovs2025gradient}. By separating device localization from anatomical image acquisition, external NMR sensing of magnetic markers offers the potential for faster 3-D tracking while reducing the RF energy deposition and acoustic noise associated with repeated tracking scans \cite{tiryaki2022magnetic,tiryaki2023magnetic}. Such feedback could support robotic guidance between anatomical image updates.

NMR-based tracking can place the field sensor either outside or on the tracked object. In \emph{off-board tracking}, external NMR probes measure field perturbations generated by device-mounted magnetic markers \cite{tiryaki2025nmr}. In \emph{on-board tracking}, object-mounted NMR probes measure externally generated magnetic fields, as demonstrated for head-motion tracking using dedicated gradient tones or native imaging gradients \cite{haeberlin2015motion,aranovitch2018prospective}. Both architectures measure only the scalar field perturbation along the scanner's main field to first order, so tracking precision depends on the source--sensor configuration. In off-board tracking, the strong MRI field saturates the magnetic markers and aligns their moments with the main field regardless of device orientation, so a single marker cannot reveal the device's orientation. These constraints motivate a common design framework for selecting receiver arrangements, marker geometries, and external source configurations to achieve the required position and device-axis precision.

We present a unified information-based framework for configuration design under the scalar-readout and field-alignment constraints of NMR magnetic tracking (Fig.~1). Through analytical modeling and numerical covariance analysis, we:
\begin{itemize}
    \item demonstrate that planar NMR tracking precision depends jointly on receiver count and array geometry;
    \item show how two spatially separated, field-aligned markers provide device-axis information for NMR tracking; and
    \item establish point-dipole source--receiver duality, unifying the analysis of on/off-board NMR tracking.
\end{itemize}

\section{Related Work}

Magnetic tracking systems acquire measurements either on the tracked device or through external receivers \cite{franz2014emt,he2025emts}. On-board approaches include conventional electromagnetic tracking and adapted planar or elongated sensor designs \cite{raab1979magnetic,cavaliere2021planar,cavaliere2022elongated}, capsule localization and closed-loop manipulation using Hall-effect sensors \cite{dinatali2013pose,taddese2018enhanced}, and wireless localization using external field gradients \cite{sharma2021wireless,sharma2023locationaware}. Electromagnetic navigation systems also support simultaneous steering and localization \cite{vonarx2024simultaneous}. Within MRI, object-mounted NMR probes have been tracked using dedicated gradient tones or native imaging gradients \cite{haeberlin2015motion,aranovitch2018prospective}. Conversely, external arrays of Hall-effect sensors or NMR probes localize device-mounted magnetic sources \cite{son2016localization,tiryaki2025nmr}. These architectures determine how device position and direction enter the measurements.

Within a chosen architecture, tracking accuracy also depends on forward-model fidelity. Finite-element-based field interpolation and learned inverse models address limitations of point-dipole modeling \cite{dinatali2016jacobian,wu2024synthetic}, while calibrated field models and online distortion mapping compensate environmental disturbances \cite{cavaliere2023calibration,sadjadi2016distortion}. These approaches improve modeling or inversion, but do not by themselves resolve information deficiencies imposed by the source--sensor configuration.

Uncertainty analysis characterizes these configuration-dependent limits. Tiryaki et al. used a local covariance bound to predict position precision for an external NMR array and experimentally evaluated tracking performance \cite{tiryaki2025nmr}. For on-board localization, Sharma et al. propagated magnetic-sensor and field-gradient errors to quantify position uncertainty \cite{sharma2021wireless}. Such analyses assess a given implementation; using uncertainty to select its configuration introduces a separate design problem.

Information-based design has been applied to sensor selection, measurement allocation, and field generation. Joshi and Boyd developed a convex-relaxation approach to covariance-based sensor selection \cite{joshi2009sensor}, and Talcoth and Rylander applied a D-optimal Fisher-information criterion to planar magnetic-sensor selection at a prescribed pose \cite{talcoth2013sensor}. Talcoth et al. subsequently optimized average and worst-case measurement allocation across a tracking domain \cite{talcoth2017allocation}, providing the closest precedent for workspace-level design. Xie et al. compared planar and split-array geometries using FIM-based 5-DoF bounds \cite{xie2026information}, while Andria et al. optimized electromagnetic field-generator geometry using field-amplitude and field-gradient objectives \cite{andria2023gradients}. The present work applies established covariance-based design principles to the coupled constraints of scalar NMR readout and field-aligned marker moments. Its contribution is a common formulation for evaluating receiver geometry, two-marker axis information, and on-board source configurations under these constraints.

\begin{figure}[t]
    \centering
    \includegraphics[width=\columnwidth]{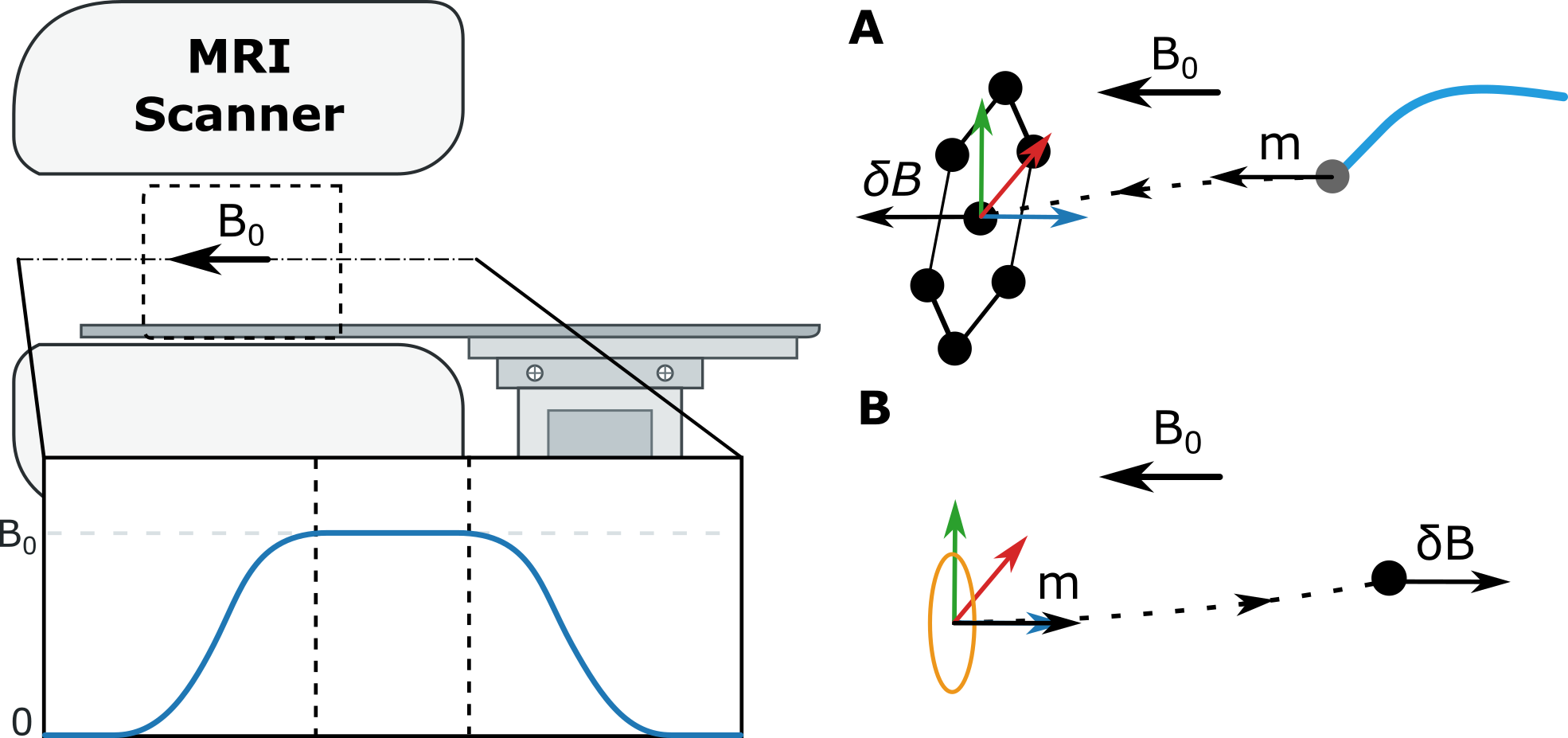}
    \vspace{-20pt}
    \caption{Magnetic-tracking architectures considered in this study.MRI setting and representative distance-dependent measurements $\delta B_i(r_i)$ from seven receivers. (A) Off-board sensing of a device-mounted, field-aligned marker $\bm{m}$ by an external receiver array. (B) On-board sensing of fields $\delta B_k$ from independently activated external sources with moments $\bm{m}_k$.}
    \label{fig:concept_figure}
    \vspace{-20pt}
\end{figure}

\section{High-Field Tracking}
\label{sec:hf_magnetic_sensing}

NMR probes have demonstrated picotesla-range sensitivity under dedicated high-field conditions \cite{gross2016dynamic}. Tracking information is nevertheless constrained by the scalar readout in both architectures and by the field-aligned magnetic moments of the off-board markers considered here.

\subsection{NMR Readout}
\label{subsec:nmr_readout}

Let $B_0=\norm{\bm{B}_0}$ and $\bm{u}_0=\bm{B}_0/B_0$ denote the magnitude and direction of the scanner's main field, and $\delta\bm{B}$ the local tracking field. The NMR resonance frequency reflects $\norm{\bm{B}_0+\delta\bm{B}}$, measured through signal phase accumulation. Since $\norm{\delta\bm{B}}\ll B_0$, the first-order approximation $\norm{\bm{B}_0+\delta\bm{B}}\approx B_0+\bm{u}_0^{\T}\delta\bm{B}$ gives the baseline-subtracted measurement $\delta B\approx\bm{u}_0^{\T}\delta\bm{B}$. In both architectures, the spatial variation of this scalar measurement provides position sensitivity.

\subsection{Magnetic Sources}
\label{subsec:magnetic_sources}

We model device-mounted magnetic markers for off-board tracking and external electromagnetic coils for on-board tracking using the point-dipole field

\begin{equation}\label{eq:external_dipole}
    \bm{B}_d(\bm{r};\bm{m}_j)
    =
    \frac{\mu_0}{4\pi\norm{\bm{r}}^3}
    \left(
        3\hat{\bm{r}}\hat{\bm{r}}^{\T}-\bm{I}_3
    \right)\bm{m}_j,
\end{equation}

where $\mu_0$ is the vacuum permeability, $\bm{m}_j$ is the magnetic moment, $\bm{r}$ points from source to observation point, and $\hat{\bm{r}}=\bm{r}/\norm{\bm{r}}$.

For the field-aligned markers considered here, $\bm{m}_j=M_j\bm{u}_0$ is independent of device direction \cite{tiryaki2025nmr}, giving the scalar field contribution

\begin{equation}\label{eq:field_aligned_dipole}
    \delta B_j(\bm{r})
    =
    \frac{\mu_0 M_j}{4\pi\norm{\bm{r}}^3}
    \left[3(\bm{u}_0^{\T}\hat{\bm{r}})^2-1\right].
\end{equation}

A single point marker therefore provides position information but no device-direction information. Our two-marker configuration provides device-axis information through the markers’ relative positions.

For a current-driven coil, the effective moment follows its area normal, with polarity set by the current \cite{phelan2022heat}. The on-board model uses independently driven coils with different moment directions.

\section{Framework}
\label{sec:framework}

We use a common measurement model to evaluate receiver geometry, marker separation, and external source configuration. Each candidate design is assessed through local position and, where applicable, device-axis uncertainty over a sampled admissible state domain.

\subsection{Measurement Model and Estimation Information}

For position tracking, the state is $\bm{q}=\bm{p}\in\R^3$. For 5-DoF tracking, $\bm{q}=(\bm{p},\bm{u})$, where the unit vector $\bm{u}$ describes the device axis with two direction degrees of freedom; axial roll is excluded.

For a candidate design $\bm{\theta}$, the measurements satisfy

\vspace{-5pt}
\begin{equation}\label{eq:generic_model}
    \bm{z}=\bm{h}(\bm{q},\bm{\theta})+\bm{n},
\end{equation}

where $\bm{\theta}$ includes receiver geometry, marker properties, and external source positions and moment directions. Measurement noise has zero mean and known, state-independent covariance $\bm{R}_n\succ0$. The weighted least-squares estimate over the admissible state domain is

\vspace{-5pt}
\begin{equation}\label{eq:wls}
    \hat{\bm{q}}
    =
    \argmin_{\bm{q}}
    \norm{\bm{z}-\bm{h}(\bm{q},\bm{\theta})}_{\bm{R}_n^{-1}}^2,
\end{equation}

where $\norm{\bm{a}}_{\bm{M}}^2=\bm{a}^{\T}\bm{M}\bm{a}$.

Small changes in the device-axis direction are represented by $\delta\bm{u}=\bm{T}_u\delta\bm{\alpha}$, where $\bm{T}_u\in\R^{3\times2}$ is an orthonormal basis perpendicular to $\bm{u}$ and $\delta\bm{\alpha}\in\R^2$ contains the two local angular perturbation coordinates, expressed in radians. For $\delta\bm{x}=(\delta\bm{p}^{\T},\delta\bm{\alpha}^{\T})^{\T}$, the linearization $\delta\bm{h}\approx\bm{H}\delta\bm{x}$ uses

\vspace{-5pt}
\begin{equation}\label{eq:state_jacobian}
    \bm{H}
    =
    \left[
        \frac{\partial\bm{h}}{\partial\bm{p}}
        \quad
        \frac{\partial\bm{h}}{\partial\bm{u}}\bm{T}_u
    \right].
\end{equation}

For position-only estimation, only the first block is retained. The local information matrix is \cite{joshi2009sensor,talcoth2017allocation}

\begin{equation}\label{eq:information_matrix}
    \bm{I}
    =
    \bm{H}^{\T}\bm{R}_n^{-1}\bm{H}.
\end{equation}

For full-column-rank $\bm{H}$, first-order propagation of measurement noise through the locally linearized weighted least-squares problem gives $\bm{\Sigma}\approx\mathcal{I}^{-1}$. Under additive Gaussian noise, $\mathcal{I}^{-1}$ is also the local Cram\'er--Rao bound for unbiased estimation; its attainability by the nonlinear estimator is not assessed here.

Position and direction uncertainty are summarized by
\begin{equation}\label{eq:criteria}
    \Ap=\sqrt{\tr\bm{\Sigma}_p},
    \qquad
    \Au=\sqrt{\tr\bm{\Sigma}_{\alpha}}.
\end{equation}

The position and direction covariance blocks are extracted after inversion of the full information matrix, retaining position--direction coupling. For position-only tracking, $\bm{\Sigma}_p=\bm{\Sigma}$. The scores $\Ap$ and $\Au$ are reported in millimeters and degrees, respectively. The latter is independent of the tangent-basis choice and describes local small-angle uncertainty.

Independent channels contribute additively, $\bm{I}=\sum_i\bm{I}_i$, whereas correlated channels require the full noise covariance. Adding an independent channel cannot worsen covariance if the existing configuration is unchanged; this guarantee does not apply when geometry, source properties, acquisition time, or noise also change.

Designs are evaluated over sampled admissible states. Local covariance is evaluated only when $\bm{W}=\bm{R}_n^{-1/2}\bm{H}\bm{S}_q\in\R^{m\times n}$ has $n$ singular values exceeding $\tau=\varepsilon_{\mathrm{mach}}\max(m,n)\sigma_{\max}(\bm{W})$. For this test, $\varepsilon_{\mathrm{mach}}$ is the double-precision machine epsilon and $\bm{S}_q=\ell\bm{I}_3$ for position-only tracking or $\bm{S}_q=\diag(\ell\bm{I}_3,\bm{I}_2)$ for 5-DoF tracking, with $\ell=1~\mathrm{mm}$. At rank-deficient states, a pseudoinverse is not interpreted as a finite full-state uncertainty bound. Uncertainty maps and summaries support design comparison: a minimum identifies the best sampled location, while average or worst-case scores characterize broader workspace performance.

Local covariance characterizes uncertainty around a given state but does not establish global uniqueness. The model symmetries described below inform the admissible workspace and the representation of the device axis; a systematic nonlocal ambiguity assessment is outside the scope of this study.
\begin{figure}[t]
    \centering
    \includegraphics[width=\columnwidth]{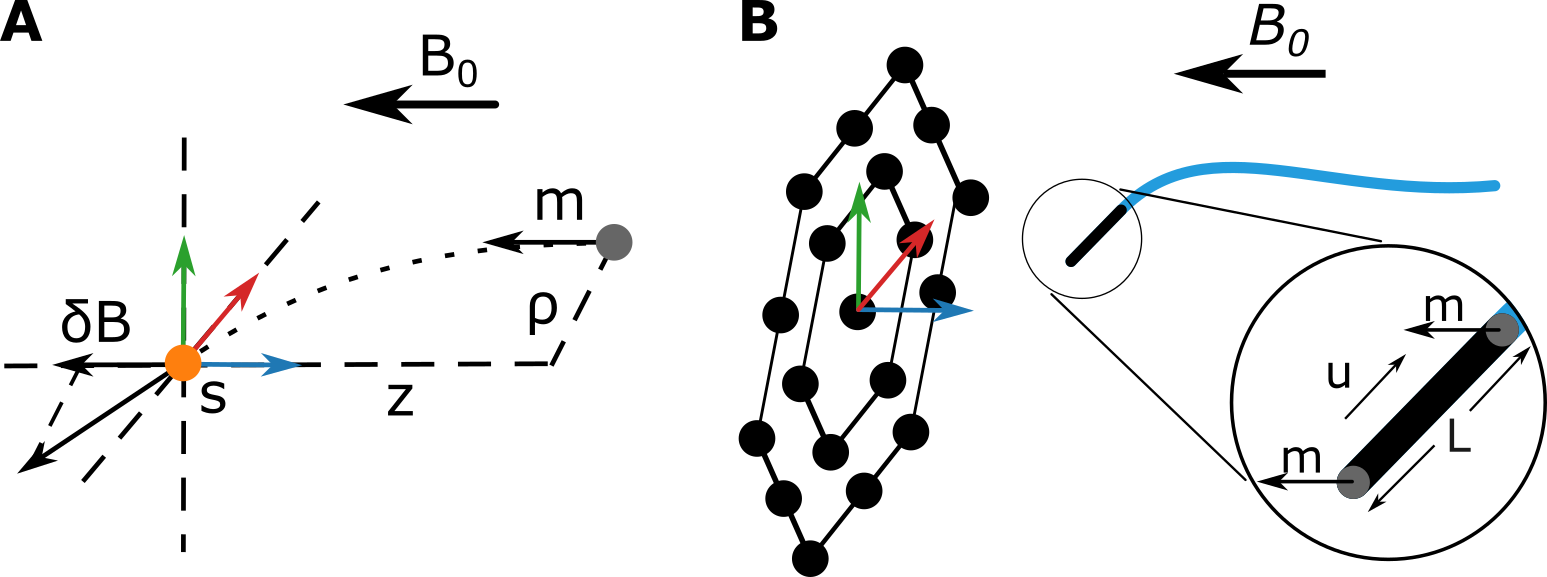}
    \vspace{-20pt}
    \caption{(A) Single marker--receiver geometry with depth $z$, lateral separation $\rho$, and local field $\delta\bm{B}$.(B) Two identical field-aligned markers separated by $L$ define the unoriented device axis $\bm{u}$. }
    \label{fig:two_marker_schematic}
    \vspace{-20pt}
\end{figure}
\subsection{Off-Board Position Sensing}
\label{subsec:offboard_position_sensing}

For external NMR receivers at positions $\bm{s}_i$, the measurement sums the scalar field contributions of the device-mounted markers:

\vspace{-10pt}
\begin{equation}\label{eq:offboard_measurement}
    h_i^{\mathrm{off}}(\bm{q})
    =
    \sum_{j=1}^{N_m}
    \delta B_j\!\left(\bm{s}_i-\bm{p}_j(\bm{q})\right),
\end{equation}

where $N_m$ is the number of device-mounted magnetic markers, $\bm{p}_j(\bm{q})$ is the position of marker $j$, and $\delta B_j$ is its scalar field contribution defined in \eqref{eq:field_aligned_dipole}.

The receiver outputs form the stacked measurement vector. For the planar configuration, receivers lie at $z=0$, with $\bm{u}_0$ normal to the plane. The resulting single-marker symmetry $h_i(x,y,z)=h_i(x,y,-z)$ makes mirrored marker positions indistinguishable, so the admissible workspace is restricted to the known half-space $z>0$.

Each scalar receiver contributes one Jacobian row. Thus, three receivers are necessary for local 3-D position estimation, but their sensitivities must also span all three position directions.

For a single marker, the field-aligned dipole model gives the receiver's total noise-normalized position sensitivity, $T(\rho,z)=\tr\bm{I}_i$, at lateral separation $\rho$ and depth $z$:

\begin{equation}\label{eq:single_sensor_information}
    T(\rho,z)
    =
    \frac{9c^2}{\sigma_B^2}
    \frac{\rho^4+4z^4}{(\rho^2+z^2)^6},
\end{equation}

where $c=\mu_0m_z/(4\pi)$, $m_z$ is the marker's signed moment along the $z$ axis, and $\sigma_B^2$ is the receiver-noise variance. At fixed $z>0$, this sensitivity decreases monotonically with lateral separation.

However, sensitivity magnitude alone does not determine array performance: each receiver supplies information along only one local direction. The full information matrix captures both the strength and directional diversity of these contributions. For fixed geometry and independent equal-variance channels, $\Ap\propto\sigma_B/|m_z|$, so uniform noise or moment scaling changes uncertainty magnitude without changing its spatial pattern.

\subsection{Two-Marker Direction Sensing}
\label{subsec:two_marker_direction}

We introduce direction information through two identical markers separated along the device axis, as illustrated in Fig.~\ref{fig:two_marker_schematic}:

\vspace{-10pt}
\begin{equation}\label{eq:two_marker_positions}
    \bm{p}_{\pm}
    =
    \bm{p}
    \pm
    \frac{L}{2}\bm{u},
\end{equation}

where $\bm{p}$ is the tracker center and $L$ is the marker separation. Their moments remain aligned with the main field, while their relative positions define the device axis. Because exchanging identical markers leaves the measurements unchanged, $\bm{u}$ and $-\bm{u}$ describe the same unoriented axis.

Let $b_i(\bm{p})$ denote one marker's contribution to receiver $i$, so that $h_i=b_i(\bm{p}_{+})+b_i(\bm{p}_{-})$. Define the gradients $\bm{g}_i^{\pm}=\nabla b_i(\bm{p}_{\pm})$, their sum $\bm{g}_i^{\Sigma}=\bm{g}_i^{+}+\bm{g}_i^{-}$, and their difference $\bm{g}_i^{\Delta}=\bm{g}_i^{+}-\bm{g}_i^{-}$. The local 5-DoF Jacobian row is

\begin{equation}\label{eq:two_marker_jacobian}
    \bm{H}_i
    =
    \left[
        (\bm{g}_i^{\Sigma})^{\T}
        \quad
        \frac{L}{2}
        (\bm{g}_i^{\Delta})^{\T}\bm{T}_u
    \right].
\end{equation}

Position sensitivity depends on the gradient sum, while direction sensitivity depends on the gradient difference. For smooth fields, $\bm{g}_i^{\Delta}=\mathcal{O}(L)$, making the direction block $\mathcal{O}(L^2)$ as $L\rightarrow0$. Closely spaced identical markers therefore provide weak direction information.

Increasing separation exposes the markers to more distinct gradients, but its effect on position and direction uncertainty depends on pose and receiver geometry. We evaluate each separation using the coupled 5-DoF covariance, obtaining $\Ap$ and $\Au$ without neglecting their interaction.

\subsection{On-Board Sequential Sensing}
\label{subsec:onboard_sequential_sensing}

A device-mounted NMR probe at position $\bm{p}$ sequentially measures the magnetic fields generated by $M$ independently activated external sources. Source $k$ is modeled as a dipole centered at $\bm{c}_k$, with moment $m_0\bm{u}_k$, where $\bm{u}_k$ is a known unit vector. Writing the dipole field as $\bm{B}_d(\bm{r};\bm{m})=\bm{D}(\bm{r})\bm{m}$, the noiseless measurement is

\begin{equation}\label{eq:onboard_field_measurement}
    h_k^{\mathrm{on}}(\bm{p})
    =
    m_0\bm{u}_0^{\T}
    \bm{D}(\bm{p}-\bm{c}_k)\bm{u}_k.
\end{equation}

The position is assumed constant throughout the sequence. The dipole tensor is symmetric and even, giving the source--receiver duality

\begin{equation}\label{eq:onboard_duality}
    \bm{u}_0^{\T}
    \bm{D}(\bm{p}-\bm{c}_k)\bm{u}_k
    =
    \bm{u}_k^{\T}
    \bm{D}(\bm{c}_k-\bm{p})\bm{u}_0.
\end{equation}

Each activated source is measurement-equivalent to an off-board scalar receiver at $\bm{c}_k$, sensitive along $\bm{u}_k$, observing a virtual dipole with moment $m_0\bm{u}_0$ at $\bm{p}$. Matched output scaling and noise covariance yield identical position Jacobians and information matrices, although the virtual receiver may not be physically realizable as an NMR probe in the same main field.

Local 3-D tracking requires at least three scalar measurements with independent position sensitivities. Co-located orthogonal sources measure the three virtual-field components, whereas co-located parallel sources repeat one projection; spatially distributed parallel sources can provide independent sensitivities. Co-located sources also retain the reflection ambiguity $\bm{h}^{\mathrm{on}}(\bm{c}+\bm{r})=\bm{h}^{\mathrm{on}}(\bm{c}-\bm{r})$ when both positions are admissible, a symmetry not generally shared by distributed sources. Source placement and moment direction must therefore be evaluated jointly.

\section{Results}
\begin{figure}[t]
    \centering
    \includegraphics[width=\columnwidth]{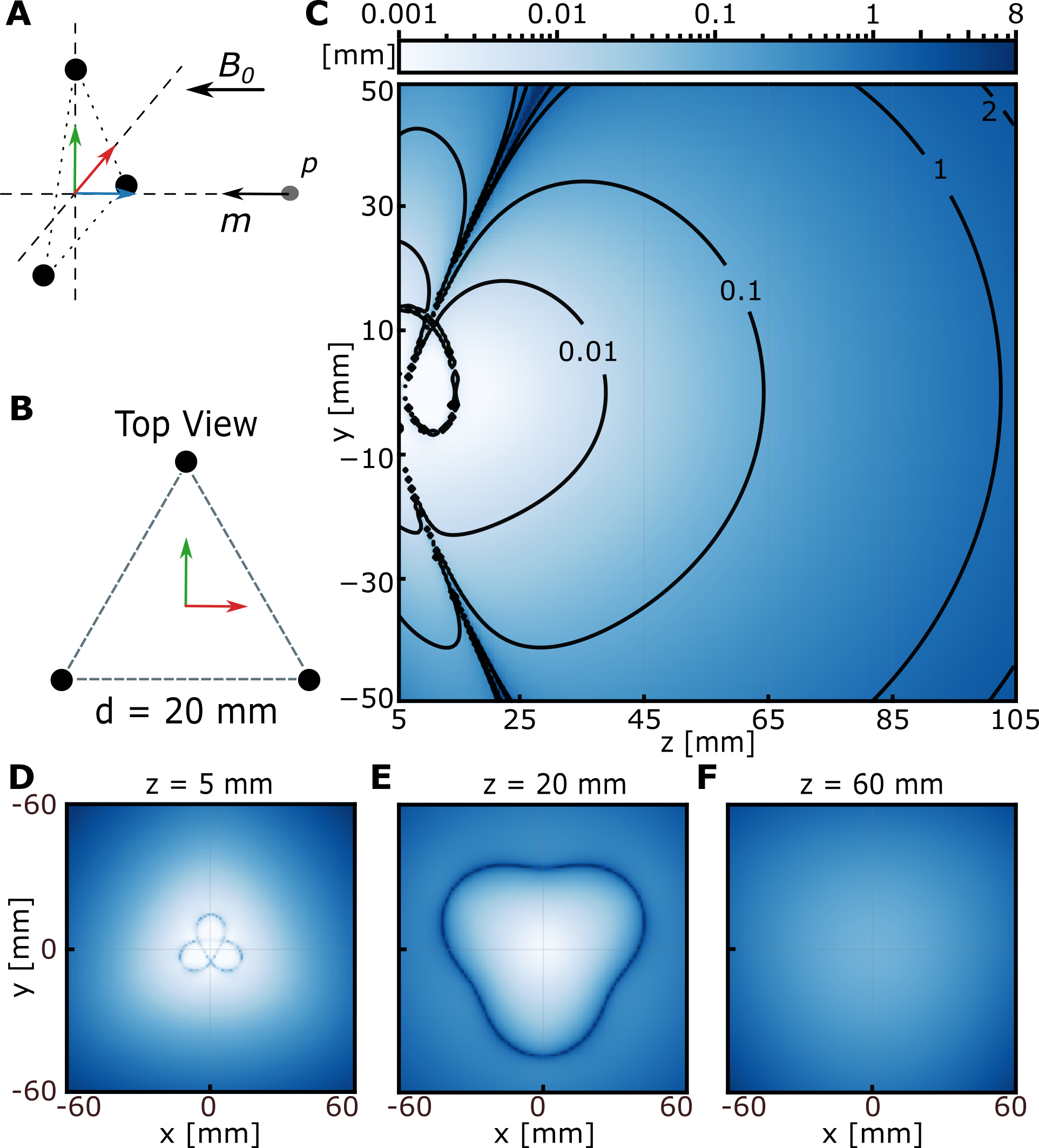}
    \vspace{-20pt}
    \caption{Local 3-D position-uncertainty score for the three-receiver planar array. (A) Measurement geometry. (B) Top view of the equilateral reference array. (C) $\Ap$ map in the $x=0$ plane. (D--F) $\Ap$ maps at $z=5$, 20, and 60 mm. All maps use a logarithmic scale with $\Ap$ reported in millimeters.}
    \label{fig:planar_workspace}
    \vspace{-20pt}
\end{figure}

We evaluate four numerical studies covering planar receiver arrays, two-marker geometry, and external source configurations for on-board tracking. The analysis assumes calibrated point-dipole models and negligible motion during sequential acquisition. Position and device-axis uncertainty are quantified by $\Ap$ and $\Au$ in \eqref{eq:criteria}, with $\Au$ evaluated only for the two-marker tracker. Throughout, the dipole-moment magnitude is $10^{-3}~\mathrm{A\,m^2}$ ($1$ emu) per marker or activated external source, and channels have independent noise with $\sigma_B=1~\mathrm{nT}$. All spatial maps represent full 3-D position estimation, including those evaluated on two-dimensional slices.

\subsection{Position Tracking with Planar Scalar Sensor Arrays}

We evaluate a single marker using an equilateral three-receiver array centered at the origin with a side length of 20 mm. This provides a minimum-count symmetric reference. The map at $x=0$ covers $y\in[-50,50]$ mm and $z\in[5,105]$ mm; maps at fixed depths of $z=5$, 20, and 60 mm cover $x,y\in[-60,60]$ mm. All maps use 0.5-mm sampling.

Figure~\ref{fig:planar_workspace}(C--F) shows strongly location-dependent position uncertainty. Near the receiver plane, $\Ap$ varies sharply around individual receivers. At greater depths, the maps become smoother but uncertainty increases, consistent with weaker field gradients.

\subsection{Receiver Count and Planar Array Geometry}

We compare triangular, square, honeycomb, and regular-polygon layouts with a minimum pairwise spacing of $d_{\min}=20~\mathrm{mm}$ and admissible receiver counts up to $N=20$. The lattice sequences start at $(0,0)$, $(d_{\min}/2,d_{\min}/2)$, and $(d_{\min},0)$, respectively, and use polar-angle lattice rings, fourfold-orbit radial shells, and sixfold-orbit radial shells; the first $N$ sites are retained and recentered at the origin. Regular polygons are generated independently as single-ring arrays with the first receiver on $+y$.

\begin{figure}[t]
    \centering
    \includegraphics[width=\columnwidth]{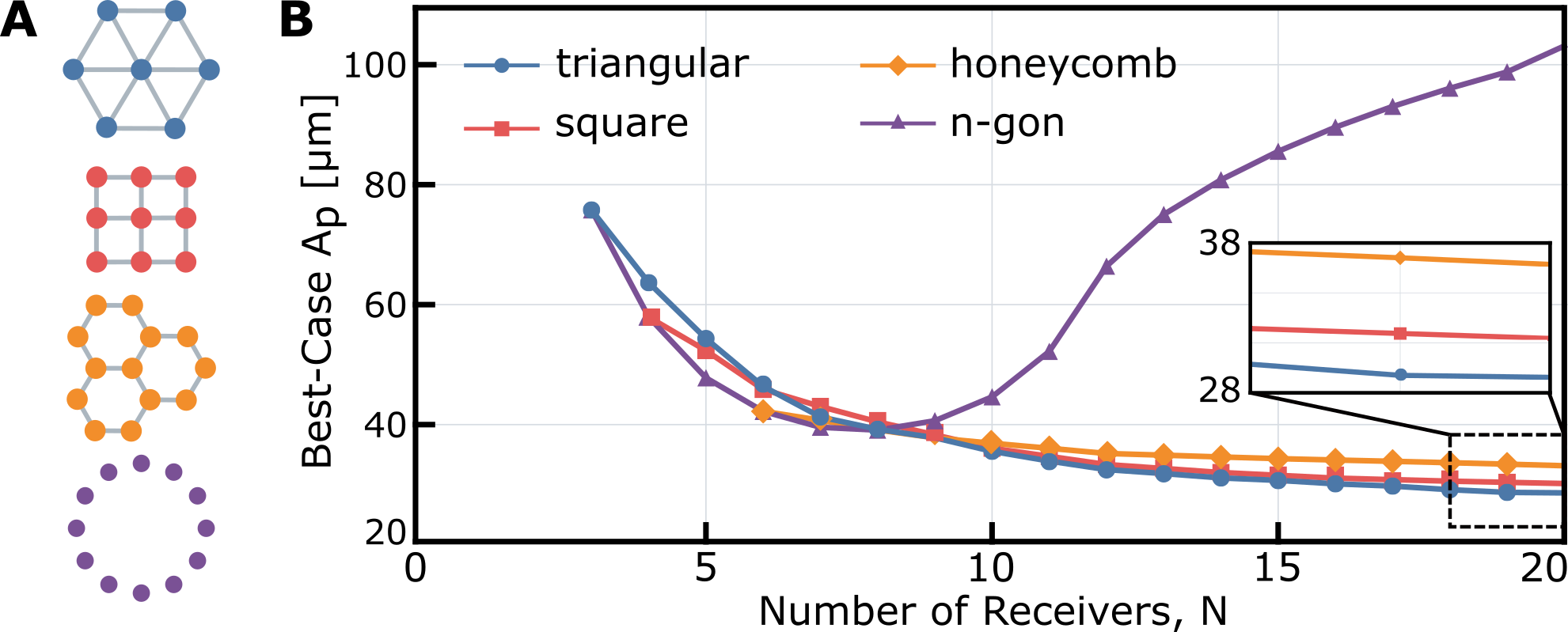}
    \vspace{-20pt}
    \caption{Planar receiver-array comparison with 20-mm minimum pairwise spacing. (A) Representative layouts from the four evaluated families. (B) Minimum sampled $\Ap$ versus receiver count $N$ at $z=60$~mm over the $100\times100$~mm$^2$ region; the inset enlarges $N=18$--20.}
    \label{fig:planar_layout_comparison}
    \vspace{-20pt}
\end{figure}

Layouts are evaluated at $z=60$ mm over $x,y\in[-50,50]$ mm with 1-mm sampling. Figure~\ref{fig:planar_layout_comparison}(B) reports the best-case local score $\min_{\bm{p}\in\mathcal{W}}\Ap(\bm{p})$ over this region. Regular polygons attain the lowest sampled minima for $4\leq N\leq8$, including ties, while the triangular family gives the lowest values for $9\leq N\leq20$. At $N=20$, the minima are 0.0287, 0.0302, 0.0331, and 0.103 mm for the triangular, square, honeycomb, and regular-polygon families, respectively.

For regular polygons, the spacing constraint sets the radius to $R_N=d_{\min}/[2\sin(\pi/N)]$. Increasing receiver count therefore expands the ring and decreases each receiver's central sensitivity according to \eqref{eq:single_sensor_information}. Layout growth and recentering also change existing receiver positions, so these curves do not represent adding channels to a fixed array. The rankings concern best-case local uncertainty: minima may occur at different positions, and no common aperture is imposed.

\subsection{Tracker Design for Rotational Information}

We evaluate two identical field-aligned markers using a planar hexagonal array of $N_s=19$ receivers with 20-mm spacing at $z=0$. Tracker centers lie at $z=60$ mm. Workspace maps use $L=20$ mm and sample $x,y\in[-60,60]$ mm at 1-mm intervals. A separate sweep fixes $\bm{p}_0=(0,0,60)~\mathrm{mm}$ and varies $L$ from 5 to 60 mm in 1-mm increments. Each score is maximized over device axes generated from direction parameters $a,b\in[-90^\circ,90^\circ]$ at $1^\circ$ intervals. Writing $a,b$ in radians and setting $\gamma=\sqrt{a^2+b^2}$, the corresponding device axis is $\bm{u}(a,b)=\left(-\frac{a\sin\gamma}{\gamma},-\frac{b\sin\gamma}{\gamma},-\cos\gamma\right)^{\T}$ for $\gamma>0$, with $\bm{u}(0,0)=-\ez$.

\begin{figure}[t]
    \centering
    \includegraphics[width=\columnwidth]{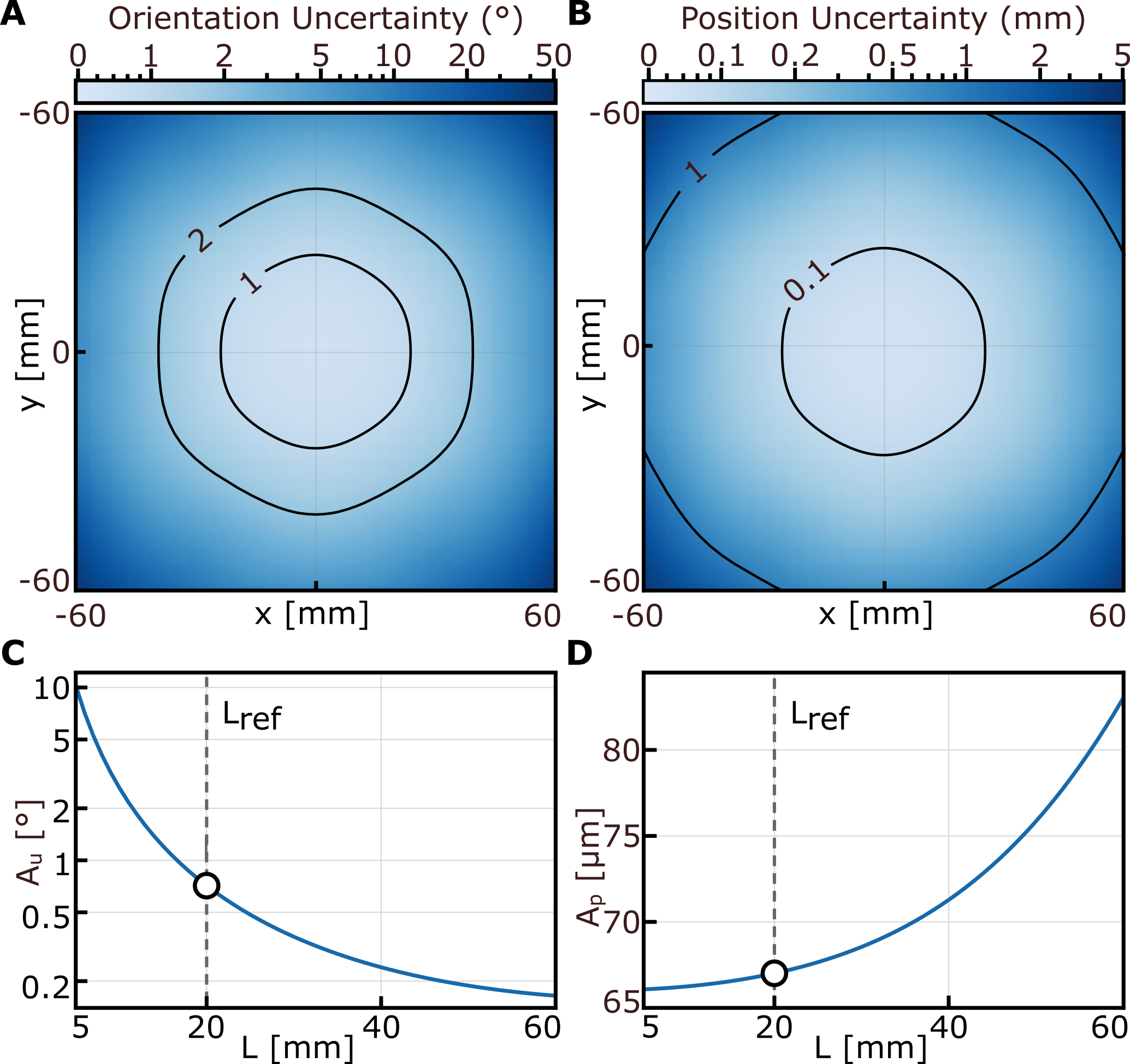}
    \vspace{-20pt}
    \caption{Worst-over-direction two-marker uncertainty. (A) $\Au$ workspace map at $z=60$ mm for $L=20$ mm; contours at $1^\circ$ and $2^\circ$. (B) Corresponding $\Ap$ map; contours at $0.1$ and $1$ mm. (C) Center-point $\Au$ versus $L$. (D) Center-point $\Ap$ versus $L$. Open circles mark $L_{\mathrm{ref}}=20$ mm; (A--C) use logarithmic scales.}
    \label{fig:two_marker_design}
    \vspace{-20pt}
\end{figure}

At $L=20$ mm, Fig.~\ref{fig:two_marker_design}(A,B) shows substantial spatial variation. The center scores are $\Au=0.713^\circ$ and $\Ap=0.0670$ mm, whereas the sampled workspace maxima reach $44.48^\circ$ and $4.47$ mm, respectively. The large angular score indicates weak local direction information rather than a validated angular-error prediction.

At the fixed center, increasing $L$ from 5 to 60 mm reduces the worst-over-direction $\Au$ monotonically from $10.28^\circ$ to $0.165^\circ$, approximately 62-fold, while $\Ap$ increases from 0.0661 to 0.0830 mm, approximately 26\% [Fig.~\ref{fig:two_marker_design}(C,D)]. Thus, the evaluated center-point sweep shows improved axis precision accompanied by higher position uncertainty and greater tracker span.

\subsection{External Source Configurations for On-Board Sensing}
\label{subsec:onboard_sensor}

We compare two configurations, each comprising three independently activated external sources. Distributed--parallel sources point along $-\ez$ and occupy the vertices of an equilateral triangle with side length 20 mm. Co-located--orthogonal sources share the origin and point along $-\bm{e}_x$, $-\bm{e}_y$, and $-\ez$. The comparison therefore changes both source placement and direction. Both configurations are evaluated on the $y=0$ plane over $x\in[-60,60]~\mathrm{mm}$ and $z\in[5,105]~\mathrm{mm}$ with 1-mm sampling. Their position Jacobians are numerically full rank at every sampled location.

\begin{figure}[t]
    \centering
    \includegraphics[width=\columnwidth]{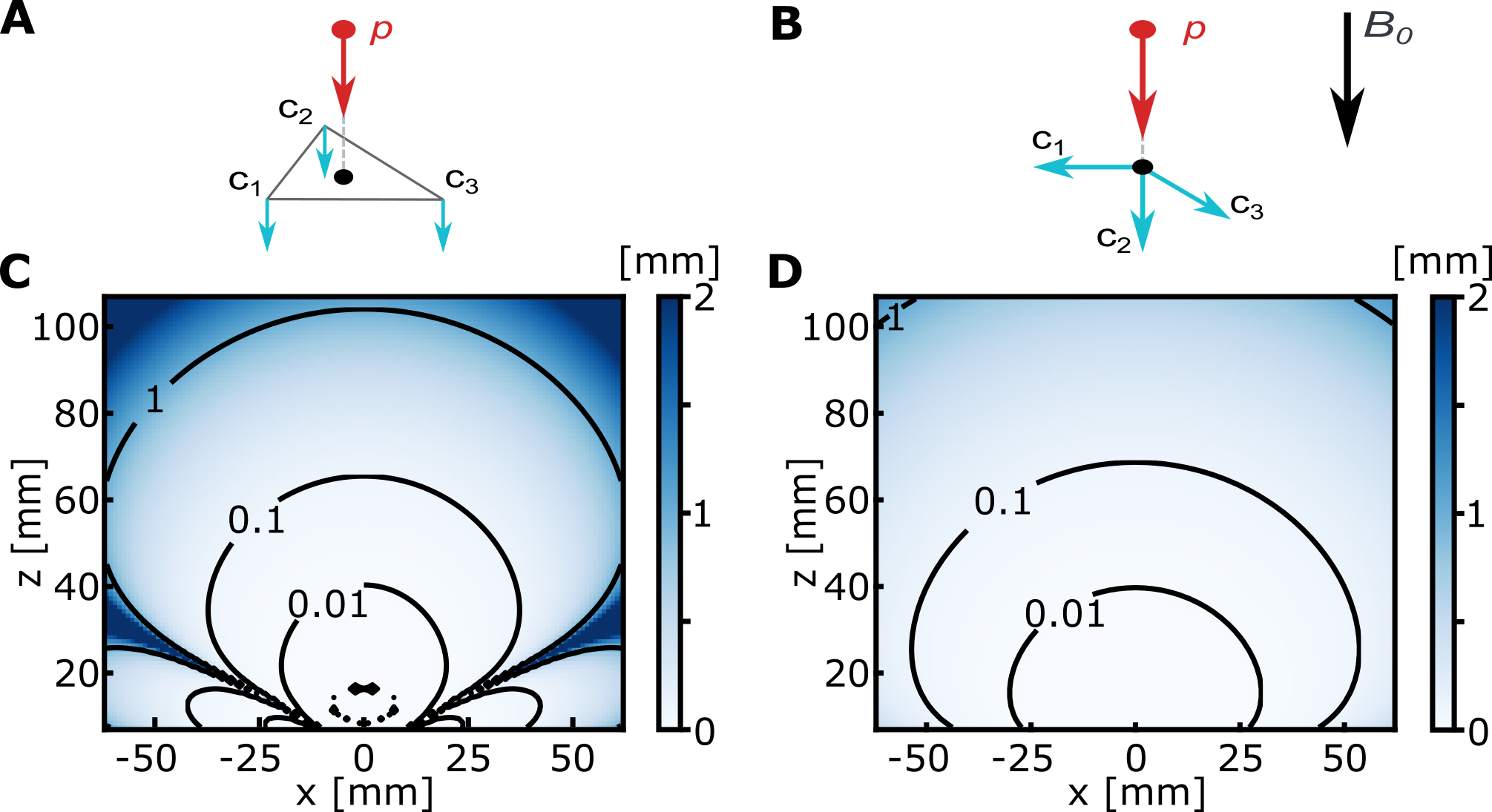}
    \vspace{-20pt}
    \caption{Sequential scalar on-board sensing with three independently activated sources. (A) Distributed--parallel configuration. (B) Co-located--orthogonal configuration. (C) Distributed--parallel $\Ap$ map in the $y=0$ plane. (D) Co-located--orthogonal $\Ap$ map. Contours mark $0.01$, $0.1$, and $1$ mm; both maps share a 0--2 mm scale, with larger values saturated.}
    \label{fig:onboard_field_mode_comparison}
    \vspace{-20pt}
\end{figure}

The co-located--orthogonal configuration reduces the sampled median $\Ap$ from $0.298$ to $0.128~\mathrm{mm}$ and the sampled maximum from $30.1$ to $1.17~\mathrm{mm}$ (Fig.~\ref{fig:onboard_field_mode_comparison}). The distributed--parallel configuration exceeds $2~\mathrm{mm}$ at $3.3\%$ of sampled locations, whereas the co-located--orthogonal configuration remains below this threshold throughout the evaluated slice. Thus, the two configurations produce markedly different uncertainty distributions despite using the same number of independently activated sources. The known central-reflection counterparts of the co-located configuration lie outside the evaluated $z>0$ half-space.

\section{Discussion}

The framework addresses two linked constraints of NMR tracking: scalar readout makes precision depend on the strength and directional diversity of spatial sensitivities, while field-aligned marker moments require spatially separated markers to encode the device axis. The studies quantify how receiver geometry, marker separation, and external source configuration affect these sensitivities within a common covariance model. The resulting design preferences remain conditional on the evaluated domain and objective: best-case array rankings need not hold for average or worst-case workspace performance~\cite{talcoth2017allocation}, and center-point separation trends need not hold throughout the workspace. The formulation accommodates 3-D receiver arrangements, although these were not evaluated here.

Magnetic moment and sensor noise provide a further design tradeoff. For fixed geometry and uniformly scaled independent channel noise, position uncertainty scales as $\Ap\propto\sigma_B/|m_z|$; the same scaling applies to $\Au$ when both marker moments are scaled equally. Thus, a proportional increase in magnetic moment compensates for higher measurement noise within the dipole model. At fixed magnetization, moment scales with marker volume, explaining why Hall-effect sensor-based systems with higher field noise require larger markers to achieve comparable precision at comparable geometry and measurement bandwidth \cite{son2016localization,tiryaki2025nmr}. Uniform noise or moment scaling preserves spatial uncertainty patterns, although differences in measurement axes and moment orientation require corresponding changes to the forward model.

Source--receiver duality unifies on/off-board tracking at the point-dipole level: matched measurement functions and noise yield identical information matrices, but not equivalent hardware or acquisition time. In particular, an equivalent receiver with arbitrary directional sensitivity may not be realizable as an NMR probe in the same main field. The framework also extends to Hall-effect sensor arrays and on-board magnetic tracking \cite{son2016localization,dinatali2013pose,taddese2018enhanced}, provided that sensor axes, device-dependent moment orientation, and noise are modeled appropriately; NMR-specific design rankings need not transfer.

The predictions rely on calibrated point-dipole models and local linearization. Model mismatch, field distortion, and parameter uncertainty may change both precision and design rankings; finite-element models and distortion compensation offer routes toward more realistic predictions \cite{dinatali2016jacobian,cavaliere2023calibration}. Large angular scores require verification beyond the small-angle approximation, and global uniqueness remains unestablished. The analysis also excludes temporal priors, process uncertainty, and motion during sequential acquisition \cite{tiryaki2020simulation}. Future work will test covariance predictions through noisy nonlinear estimation, incorporate model uncertainty and temporal dynamics, and validate candidate configurations experimentally on moving robotic systems.

\bibliographystyle{IEEEtran}
\bibliography{references}
\end{document}